\documentclass{article}
\usepackage{ijcai21}

\usepackage{times}
\usepackage{soul}
\usepackage{url}
\usepackage[hidelinks]{hyperref}
\usepackage[utf8]{inputenc}
\usepackage[small]{caption}
\usepackage{graphicx}
\usepackage{amsmath}
\usepackage{amsthm}
\usepackage{booktabs}
\usepackage{algorithm}
\usepackage{algorithmic}
\usepackage{amssymb}

\usepackage{color}
\usepackage{multirow}
\usepackage[table]{xcolor}
\usepackage{arydshln}
\usepackage{threeparttable}
\usepackage{colortbl}
\usepackage{enumitem}
\usepackage{siunitx}
\usepackage{bm}
\usepackage{hhline}
\usepackage{array}
\usepackage[table]{xcolor}
\usepackage{caption}
\usepackage{dblfloatfix}

\usepackage{multibib}
\newcites{Appendix}{Supplementary References}

\definecolor{mygray}{gray}{.9}
\definecolor{ggray}{RGB}{127,127,127}
\definecolor{reda}{RGB}{192,0,0}
\definecolor{redb}{RGB}{217,148,143}
\definecolor{myyellow}{RGB}{190,144,0}
\definecolor{mygreen}{RGB}{80,100,40}
\definecolor{myblue}{RGB}{30,90,100}

\makeatletter
\newcommand{\thickhline}{%
    \noalign {\ifnum 0=`}\fi \hrule height 1pt
    \futurelet \reserved@a \@xhline
}
\makeatother

\newcommand{\etal}{\textit{et al}.}
\newcommand{\ie}{\textit{i}.\textit{e}.}
\newcommand{\eg}{\textit{e}.\textit{g}.}
\newcommand{\etc}{\textit{etc}.}
\newcommand{\wrt}{\textit{w}.\textit{r}.\textit{t}.}
\newcommand{\cf}{\textit{c}.\textit{f}.}

\usepackage[page]{appendix} 

\title{Information Bottleneck Approach to Spatial Attention Learning}

\author{
Qiuxia Lai$^1$
\and
Yu Li$^1$\and
Ailing Zeng$^1$\and
Minhao Liu$^1$\and
Hanqiu Sun$^2$\And
Qiang Xu$^1$
\affiliations
$^1$The Chinese University of Hong Kong\\
$^2$University of Electronic Science and Technology of China\\
\emails
\{qxlai, yuli, alzeng, mhliu, qxu\}@cse.cuhk.edu.hk,
}

\begin{document}

\maketitle

\begin{abstract}
  The selective visual attention mechanism in the human visual system (HVS) restricts the amount of information to reach visual awareness for perceiving natural scenes, allowing near real-time information processing with limited computational capacity~\cite{koch1987shifts}. This kind of selectivity acts as an `Information Bottleneck (IB)', which seeks a trade-off between information compression and predictive accuracy. 
  However, such information constraints are rarely explored in the attention mechanism for deep neural networks (DNNs). 
  In this paper, we propose an IB-inspired spatial attention module for DNN structures built for visual recognition. 
  The module takes as input an intermediate representation of the input image, and outputs a variational 2D attention map that minimizes the mutual information (MI) between the attention-modulated representation and the input, while maximizing the MI between the attention-modulated representation and the task label. 
  To further restrict the information bypassed by the attention map, 
  we quantize the continuous attention scores to a set of learnable anchor values during training.
  Extensive experiments show that the proposed IB-inspired spatial attention mechanism can yield attention maps that neatly highlight the regions of interest while suppressing backgrounds, and bootstrap standard DNN structures for visual recognition tasks (\eg, image classification, fine-grained recognition, cross-domain classification). 
  The attention maps are interpretable for the decision making of the DNNs as verified in the experiments.
  Our code is available at \href{https://github.com/ashleylqx/AIB.git}{this https URL}.
\end{abstract}

\section{Introduction}

Human beings can process vast amounts of visual information in parallel through visual system~\cite{koch2006much} because the attention mechanism can selectively attend to the most informative parts of visual stimuli rather than the whole scene~\cite{eriksen1972temporal}. 
A recent trend is to incorporate attention mechanisms into deep neural networks (DNNs), to focus on task-relevant parts of the input automatically. 
Attention mechanism has benefited sequence modeling tasks as well as a wide range of computer vision tasks to boost the performance and improve the interpretability. 

The attention modules in CNNs can be broadly categorized into \textit{channel-wise attention} and \textit{spatial attention}, which learns channel-wise~\cite{hu2018squeeze} and spatially-aware attention scores~\cite{simonyan2013deep} for modulating the feature maps, respectively. 
As channel-wise attention would inevitably lose spatial information essential for localizing the important parts, in this paper, we focus on the spatial attention mechanism. 
There are \textit{query-based} and \textit{module-based} spatial attention learning methods.
Query-based attention, or `self-attention', generates the attention scores based on the similarity/compatibility between the query and the key content. Though having facilitated various computer vision tasks, such dense relation measurements would lead to heavy computational overheads~\cite{han2020survey}, which significantly hinders its application scenario.
Module-based attention directly outputs an attention map using a learnable network module that takes as input an image/feature. Such an end-to-end inference structure is more efficient than query-based attention, and has been shown to be effective for various computer vision tasks.
Existing spatial attention mechanisms trained for certain tasks typically generate the attention maps by considering the contextual relations of the inputs. Although the attention maps are beneficial to the tasks, the attention learning process fails to consider the inherent information constraints of the attention mechanism in HVS, 
\ie, to ensure that the information bypassed by the attention maps is of minimal redundancy, and meanwhile being sufficient for the task.

To explicitly incorporate the information constraints of the attention mechanism in HVS into attention learning in DNNs, in this paper, we propose an end-to-end trainable spatial attention mechanism inspired by the `Information Bottleneck (IB)' theory. 
The whole framework is derived from an information-theoretic argument based on the IB principle. The resulted variational attention maps can effectively filter out task-irrelevant information, which reduces the overload of information processing while maintaining the performance. 
To further restrict the information bypassed by the attention map, an adaptive quantization module is incorporated to round the attention scores to the nearest anchor value. 
In this way, previous continuous attention values are replaced by a finite number of anchor values, which further compress the information filtered by the attention maps.
To quantitatively compare the interpretability of the proposed attention mechanism with others, 
we start from the definition of interpretability on model decision making, and measure the interpretability in a general statistical sense based on the attention consistency between the original and the modified samples that do not alter model decisions.

In summary, our contributions are three-fold:
\begin{itemize}
  \item We propose an IB-inspired spatial attention mechanism for visual recognition, which yields variational attention maps that minimize the MI between the attention-modulated representation and the input while maximizing the MI between the attention-modulated representation and the task label.  
  
  \item To further filter out irrelevant information, we design a quantization module to round the continuous attention scores to several learnable anchor values during training. 
  
  \item The proposed attention mechanism is shown to be more interpretable for the decision making of the DNNs compared with other spatial attention models.
\end{itemize}

Extensive experiments validate the theoretical intuitions behind the proposed IB-inspired spatial attention mechanism, and show improved performances and interpretability for visual recognition tasks.

\section{Related Work}
\label{section:rw}

\subsection{Spatial Attention Mechanism in DNNs}

Attention mechanisms enjoy great success in sequence modeling tasks such as machine translation~\cite{bahdanau2015neural}, speech recognition~\cite{chorowski2015attention} and image captioning~\cite{xu2015show}. Recently, they are also shown to be beneficial to a wide range of computer vision tasks to boost the performance and improve the interpretability of general CNNs. 
The attention modules in CNNs fall into two broad categories, namely channel-wise attention and spatial attention. The former learns channel-wise attention scores to modulate the feature maps by reweighing the channels~\cite{hu2018squeeze}. The latter learns a spatial probabilistic map over the input to enhance/suppress each 2D location according to its relative importance \wrt the target task~\cite{simonyan2013deep}. 
In this section, we focus on spatial attention modules.

\noindent\textbf{Query-based/ Self-attention.}
Originated from query-based tasks~\cite{bahdanau2015neural,xu2015show}, this kind of attention is generated by measuring the similarity/compatibility between the query and the key content. 
Seo~\etal~\shortcite{seo2018progressive} use a one-hot encoding of the label to query the image and generate progressive attention for attribute prediction.
Jetley~\etal~\shortcite{jetley2018learn} utilize the learned global representation of the input image as a query and calculate the compatibility with local representation from each 2D spatial location. 
Hu~\etal~\shortcite{hu2019local} adaptively determines the aggregation weights by considering the compositional relationship of visual elements in local areas. 
Query-based attention considers dense relations in the space, which can improve the discriminative ability of CNNs. However, the non-negligible computational overhead limits its usage to low-dimensional inputs, and typically requires to downsample the original images significantly.

\noindent\textbf{Module-based attention.}
The spatial attention map can also be directly learned using a softmax-/sigmoid-based network module that takes an image/feature as input and outputs an attention map. Being effective and efficient, this kind of attention module has been widely used in computer vision tasks such as action recognition~\cite{sharma2015action} and image classification~\cite{woo2018cbam}.
Our proposed spatial attention module belongs to this line of research, and we focus on improving the performance of visual recognition over baselines without attention mechanisms.

Previous spatial attention learning works focus on the relations among non-local or local contexts to measure the relative importance of each location, and do not consider the information in the feature filtered by the attention maps. We instead take inspiration from the IB theory~\cite{tishby1999information} which maintains a good trade-off between information compression and prediction accuracy, and propose to learn spatial attention that minimizes the MI between the masked feature and the input, while maximizing the MI between the masked feature and the task label. 
Such information constraints could also help remove the redundant information from the input features compared with conventional relative importance learning mechanisms.

\begin{figure*}
\begin{minipage}{\textwidth}
\begin{minipage}[t]{0.245\textwidth}
  \centering
      \includegraphics[width=0.4\textwidth]{graph_uni}
  \captionsetup{font=small}

    \makeatletter\def\@captype{figure}\makeatother\caption{\textbf{Graphical model} of the probabilistic neural network with IB-inspired spatial attention mechanism (\S\ref{sec:theory}).
    \label{fig:graph}}
  \end{minipage}
\hfill
\begin{minipage}[t]{0.74\textwidth}
 \centering
      \includegraphics[width=1\textwidth]{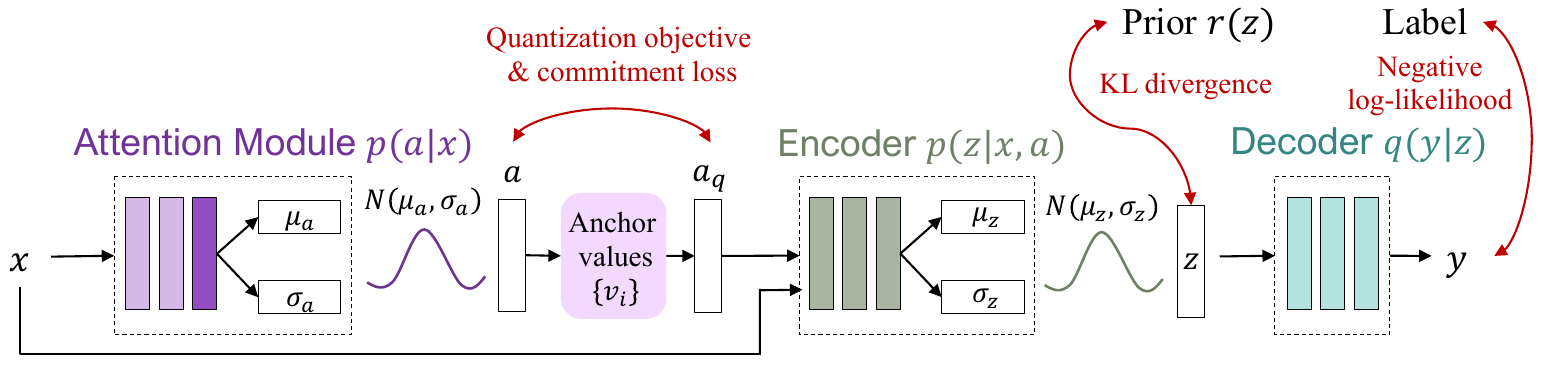}

  \captionsetup{font=small}
    \makeatletter\def\@captype{figure}\makeatother\caption{\textbf{Framework of the IB-inspired spatial attention mechanism for visual recognition.} 
    The input $x$ is passed through an attention module to produce a continuous variational attention map $a$, which is quantized to a discrete attention map $a_q$ using a set of learnable anchor values $v_i$.
    Then, $a_q$ and $x$ are encoded to a latent vector $z$, and decoded to a prediction $y$. 
    Loss function in Eq.~(\ref{eq:loss_func}).
    See \S\ref{sec:theory} and \S\ref{sec:quantization}.   
    \label{fig:framework}} 
 \end{minipage}
\end{minipage}
  \vspace{-16pt} 
\end{figure*}

\subsection{IB-Inspired Mask Generation}

IB theory has been explored in tasks such as representation learning~\cite{alemi2017deep,achille2018information} and domain adaptation~\cite{du2020learning}. 
Here, we focus on IB-inspired mask generation methods, 
which yield additive~\cite{schulz2019restricting} or multiplicative masks~\cite{achille2018information,taghanaki2019infomask,zhmoginov2019information} to restrict the information that flows to the successive layers.

Information Dropout~\cite{achille2018information} multiplies the layer feature with a learnable noise mask to control the flow of information. This is equivalent to optimizing a modified cost function that approximates the IB Lagrangian of~\cite{tishby1999information}, allowing the learning of a sufficient, minimal and invariant representation for classification tasks.
InfoMask~\cite{taghanaki2019infomask} filters out irrelevant background signals using the masks optimized on IB theory, which improves the accuracy of localizing chest disease.
Zhmoginov~\etal~\shortcite{zhmoginov2019information} generates IB-inspired Boolean attention masks for image classification models to completely prevent the masked-out pixels from propagating any information to the model output.
Schulz~\etal~\shortcite{schulz2019restricting} adopts the IB concept for interpreting the decision-making of a pre-trained neural network, by adding noise to intermediate activation maps to restrict and quantify the flow of information. The intensity of the noise is optimized to minimize the information flow while maximizing the classification score.

In this paper, we propose an IB-inspired spatial attention mechanism based on a new variational approximation of the IB principle derived from a neural network incorporated with attention module ($\mathcal{L}_{\text{AttVIB}}$ in Eq.~(\ref{eq:IB_two})). 
Compared with the above works, our optimization objective is not derived from the original IB Lagrangian of~\cite{tishby1999information}, thus the resulted attention mechanism is parameterized with different network architectures. 
Besides, we focus on learning continuous spatial attention instead of noise masks~\cite{achille2018information,schulz2019restricting}, which is also different from the Boolean attention in~\cite{zhmoginov2019information}.

\section{Methodology}
\label{sec:method}

\subsection{Overview}
\label{sec:overview}

We derived the framework of the proposed spatial attention mechanism theoretically from the IB principle on a neural network incorporated with an attention module. 
Assuming that random variables $X, Y, A$ and $Z$ follow the joint conditional distribution of the model shown in Fig.~\ref{fig:graph}. Here, $X, Y, A$ and $Z$ denote the input, output, spatial attention map, and the latent representation obtained from $X$ and $A$, respectively. 
We drive the proposed spatial attention framework based on the joint distribution and the IB principle. 
The resulted framework is shown in Fig.~\ref{fig:framework}, which consists of an attention module $p(a|x)$, a set of anchor values $\{v_i\}$ for quantization, an encoder $p(z|x,a)$, and a decoder $q(y|z)$.

The input $x$ is first passed through the attention module $p(a|x)$ to produce a continuous variational attention map $a$, which is quantized to a discrete attention map $a_q$ using a set of learnable anchor values $v_i$. Then, $a_q$ and $x$ are encoded to a latent vector $z$, and decoded to a prediction $y$. 
The whole framework is trained end-to-end using the loss function defined in Eq.~(\ref{eq:loss_func}).
The derivation of the framework is shown in~\S\ref{sec:theory}, and more details are provided in the supplementary.
The attention score quantization process is shown in~\S\ref{sec:quantization}.

\subsection{IB-inspired Spatial Attention Mechanism}
\label{sec:theory}

We introduce the IB principle~\cite{tishby1999information} to learn the spatial attention that minimizes the MI between the masked representation and the input, while maximizing the MI between the masked representation and the class label. 
Different from~\cite{alemi2017deep} for a standard learning framework, we derive new variational bounds of MI within an \textit{attentive framework} for visual recognition, which lead to an IB-inspired spatial attention mechanism. 

Let the random variables $X, Y, A$ and $Z$ denote the input, output, spatial attention map, and the latent representation obtained from $X$ and $A$, respectively. The MI between the latent feature $Z$ and its output $Y$ is defined as:
\begin{equation}\label{eq:ZY_lb}
I(Z; Y)= \int p(y, z) \log \frac{p(y|z)}{p(y)} dydz.
\end{equation}
We introduce $q(y|z)$ to be a variational approximation of the intractable $p(y|z)$. Since the Kullback Leibler (KL) divergence is always non-negative, we have:
\begin{equation}\label{eq:KL_nonneg}
\begin{split}
&D_{\text{KL}}[p(y|z)||q(y|z)]=\int p(y|z)\log \frac{p(y|z)}{q(y|z)} dy \ge 0 \\
        &\Rightarrow \int p(y|z)\log p(y|z) dy \ge \int p(y|z)\log q(y|z) dy, 
\end{split}
\end{equation}
which leads to
\begin{equation}\label{eq:ZY_lb}
I(Z; Y)\ge \int p(y, z) \log q(y|z)dydz + H(Y),
\end{equation}
where $H(Y)\!=\!-\int p(y) \log p(y) dy$ is the entropy of $Y$, which is independent of the optimization procedure thus can be ignored. 
By leveraging the fact that $p(y, z)\!=\!\int p(x, y, z, a)dxda\!=\!\int p(x)p(a|x)p(y|x, a)p(z|x, a)dxda$ (see Fig.~\ref{fig:graph}), and ignoring $H(Y)$, the new variational lower bound is as follows:
\begin{equation}\label{eq:ZY_lb_two}
\resizebox{.91\linewidth}{!}{$
I(Z; Y)\!\ge\!\int p(x, y) p(a|x) p(z|x, a)\!\log q(y|z) dxdadydz, 
$}
\end{equation}
where $p(a|x)$ is the attention module, $p(z|x, a)$ is the encoder, and $q(y|z)$ is the decoder.
$I(Z; Y)$ can thus be maximized by maximizing its variational lower bound.

Next, to minimize the MI between the attention-modulated representation and the input, we consider $I(Z; X, A)$, and obtain the following upper bound:
\begin{equation}\label{eq:ZXA_ub}
\resizebox{.91\linewidth}{!}{$
I(Z; X, A)\!\le\!\int p(x)p(a|x)p(z|x, a)\!\log \frac{p(z|x, a)}{r(z)} dxdadz,
$}
\end{equation}
where $r(z)$ is a prior distribution of $z$. 
In our experiments, we use 
a spherical Gaussian $\mathcal{N}(z|0,I)$ as the prior $r(z)$.

By combining Eq.~(\ref{eq:ZY_lb_two}) and~(\ref{eq:ZXA_ub}), we obtain the lower bound of the attentive variational information bottleneck:
\begin{equation}\label{eq:IB_two}
\begin{split}
\mathcal{L}_{\text{AttVIB}} &\equiv I(Z; Y) - \beta I(Z;X, A) \\
= \int &p(x, y) p(a|x) p(z|x, a) \log q(y|z) dxdadydz \\
  - \beta &\int p(x)p(a|x)p(z|x, a) \log \frac{p(z|x, a)}{r(z)} dxdadz,
\end{split}
\end{equation}
which offers an IB-inspired spatial attention mechanism for visual recognition. Here, $\beta\!>\!0$ controls the tradeoff between information compression and prediction accuracy.

We approximate $p(x)$, $p(x, y)$ with empirical distribution $p(x)\!=\!\frac{1}{N}\sum_{n=1}^{N}\delta_{x_n}(x)$, $p(x, y)\!=\!\frac{1}{N}\sum_{n=1}^{N}\delta_{x_n}(x)\delta_{y_n}(y)$ following~\cite{alemi2017deep}, where $N$ is the number of training samples, $x_n$ and $(x_n, y_n)$ are samples drawn from data distribution $p(x)$ and $p(x, y)$, respectively. 
The approximated lower bound $\mathcal{L}_{\text{AttVIB}}$ can thus be written as:
\begin{equation}\label{eq:L_IB}
\resizebox{.91\linewidth}{!}{$
\begin{split}
\widetilde{\mathcal{L}}_{\text{AttVIB}} = &\frac{1}{N}\sum_{n=1}^{N}  \{\int p(a|x_n) p(z|x_n, a) \log q(y_n|z) dadz \\
        -\beta &\int p(a|x_n)p(z|x_n, a) \log \frac{p(z|x_n, a)}{r(z)} dadz\}.
\end{split}
$}
\end{equation}
Similar to~\cite{alemi2017deep}, we suppose to use the attention module of the form $p(a|x)\!=\!\mathcal{N}(a| g_e^{\mu}(x), g_e^{\Sigma}(x))$, and the encoder of the form $p(z| x, a)\!=\!\mathcal{N}(z| f_e^{\mu}(x, a), f_e^{\Sigma}(x, a))$, where $g_e$ and $f_e$ are network modules. 
To enable back-propagation, we use the reparametrization trick~\cite{kingma2014auto}, and write $p(a|x)da\!=\!p(\varepsilon)d\varepsilon$ and $p(z|x, a)dz\!=\!p(\epsilon)d\epsilon$, where $a\!=\!g(x, \varepsilon)$ and $z\!=\!f(x, a, \epsilon)$ are deterministic functions of $x$ and the Gaussian random variables $\varepsilon$ and $\epsilon$. 
The loss function is:
\begin{equation}\label{eq:L_IB_RP}
\resizebox{.91\linewidth}{!}{$
\begin{split}
\mathcal{L}\!=\!-&\frac{1}{N}\sum_{n=1}^{N}\!\{ \mathbb{E}_{\epsilon\sim p(\epsilon)}[\mathbb{E}_{\varepsilon\sim p(\varepsilon)}[\log q(y_n|f(x_n, g(x_n, \varepsilon), \epsilon))]] \\
+ &\beta \mathbb{E}_{\varepsilon\sim p(\varepsilon)}[D_{\text{KL}}[p(z|x_n, g(x_n, \varepsilon)) \parallel r(z)]]\}.
\end{split}
$}
\end{equation}
The first term of the loss function is the negative log-likelihood of the prediction, where the label $y_n$ of $x_n$ is predicted from the latent encoding $z$, and $z$ is generated from $x_n$ and its attention map $a\!=\!g(x_n, \varepsilon)$. Minimizing this term leads to maximal prediction accuracy. 
The second term is the KL divergence between distributions of latent encoding $z$ and the prior $r(z)$, the minimization of which enables the model to learn an IB-inspired spatial attention mechanism for visual recognition.
This is different from the regular IB principle~\cite{tishby1999information,alemi2017deep} which is only for the representation learning without attention module.

\begin{table*}[t]
\begin{center}

\begin{threeparttable}
    \resizebox{0.9\textwidth}{!}{
    \renewcommand\arraystretch{0.95}
    \begin{tabular}{l|| c c| c c| c c}
    \hline\thickhline
    &\multicolumn{2}{c|}{Image Class.}
                           &\multicolumn{2}{c|}{Fine-grained Recog.}
                           &\multicolumn{2}{c}{Cross-domain Class.} \\
    \cline{2-7}                       
    \multirow{-2}{*}{Model} &CIFAR-10 &CIFAR-100 
                            &CUB-200-2011 &SVHN 
                            &STL10-train &STL10-test \\

    \hline
    \multicolumn{3}{l}{-- Existing architectures --} \\
    VGG~\cite{simonyan2014very}     &7.77 &30.62 
                                    &34.64 &4.27 
                                    &54.66 &55.09\\
    VGG-GAP~\cite{zhou2016learning} &9.87 &31.77 
                                    &29.50 &5.84
                                    &56.76 &57.24\\
    VGG-PAN~\cite{seo2018progressive}  &6.29 &24.35 
                                       &31.46 &8.02 
                                       &52.50  &52.79\\
    
    VGG-DVIB~\cite{alemi2017deep}   &~~4.64$^*$ &~~22.88$^*$ 
                                    &~~\textbf{23.94}$^*$ &~~3.28$^*$ 
                                    &~~\textbf{51.40}$^*$ &~~\textbf{51.60}$^*$ \\                                                                  
                                      
    WRN~\cite{zagoruyko2016wide}   &\textbf{4.00} &\textbf{19.25} 
                                   &26.50 &– 
                                   &– &– \\

    \hline
    \multicolumn{3}{l}{-- Architectures with attention --} \\
    VGG-att2~\cite{jetley2018learn}   &5.23 &23.19 
                                      &26.80 &3.74 
                                      &51.24  &51.71\\
    VGG-att3~\cite{jetley2018learn} &6.34 &22.97  
                                    &26.95 &3.52 
                                    &51.58 &51.68\\
    WRN-ABN~\cite{fukui2019attention} &~~3.92$^*$  &18.12 
                                      &– &~~2.88$^*$  
                                      &~~50.90$^*$ &~~51.24$^*$ \\

    VGG-aib (ours)              &4.28 &21.56 
                                &23.73 &3.24 
                                &50.64 &51.24\\

    VGG-aib-qt (ours)           &\textbf{4.10} &\textbf{20.87} 
                                &\textbf{21.83} &\textbf{3.07} 
                                &\textbf{50.44} &\textbf{51.16}\\

    WRN-aib (ours)              &3.60 &17.82 
                                &17.26 &2.76    
                                &\textbf{50.08} &50.84\\
    WRN-aib-qt (ours)           &\textbf{3.43} &\textbf{17.64} 
                                &\textbf{15.50} &\textbf{2.69}
                                &50.34 &\textbf{50.49}\\

    \hline
    \end{tabular}
}
\end{threeparttable}
\end{center}
\caption{Top-1 error for image classification~(\S\ref{sec:image_cls}), fine-grained recognition~(\S\ref{sec:fg_cls}), and cross-dataset classification~(\S\ref{sec:cd_cls}).  $^*$ denotes re-implementation or re-training. Other values are from the original paper. Best values of different backbones are in \textbf{bold}.
 \label{table:visual_recognition}}
\end{table*}

\subsection{Attention Score Quantization} 
\label{sec:quantization}

We define the continuous attention space as $A\!\in\!\mathbb{R}^{W_a\!\times\!H_a}$, and the quantized attention space as $A_q\!\in\!\mathbb{R}^{W_a\!\times\!H_a}$, where $W_a, H_a$ are the width and height of the attention map, respectively. 
As shown in Fig.~\ref{fig:framework}, the input $x$ is passed through an attention module to produce a continuous variational attention map $a$, which is mapped to a discrete attention map $a_q$ through a nearest neighbour look-up among a set of learnable anchor values $\{v_i\!\in\!\mathbb{R}\}_{i=1}^Q$, which is given by:
\begin{equation}\label{eq:qt_att}
a_q^{(w,h)}=v_k,\;k=\arg\min_j \Vert a^{(w,h)}-v_j \Vert_2,
\end{equation}
where $w\!=\!1\dots W_a, h\!=\!1\dots H_a$ are spatial indices. In this way, each score $a^{(w,h)}$ in the continuous attention map is mapped to the 1-of-$Q$ anchor value. 
The quantized attention map $a_q$ and the input $x$ are then encoded into a latent representation $z\!\in\!\mathbb{R}^K$, where $K$ is the dimension of the latent space. Finally, $z$ is mapped to the prediction probabilities $y\!\in\!\mathbb{R}^C$, and $C$ is the number of classes. 
The complete model parameters include the parameters of the attention module, encoder, decoder, and the anchor values $\{v_i\!\in\!\mathbb{R}\}_{i=1}^Q$.  

As the $\arg\min$ operation in Eq.~(\ref{eq:qt_att}) is not differentiable, we resort to the straight-through estimator~\cite{bengio2013estimating} and approximate the gradient of $a_q$ using the gradients of $a$. Though simple, this estimator worked well for the experiments in this paper.
To be concrete, in the forward process the quantized attention map $a_q$ is passed to the encoder, and during the backwards computation, the gradient of $a$ is passed to the attention module unaltered. Such a gradient approximation makes sense because $a_q$ and $a$ share the same $W_a\!\times\!H_a$ dimensional space, and the gradient of $a$ can provide useful information on how the attention module could change its output to minimize the loss function defined in Eq.~(\ref{eq:L_IB_RP}). 

The overall loss function is thus defined as in Eq.~(\ref{eq:loss_func}), which extends Eq.~(\ref{eq:L_IB_RP}) with two terms, namely a quantization objective $\Vert \text{sg}[g(x_n, \varepsilon)]\!-\!a_q^\varepsilon\Vert_2^2$ weighted by $\lambda_q$, and a commitment term $\Vert g(x_n, \varepsilon)\!-\!\text{sg}[a_q^\varepsilon] \Vert_2^2$ weighted by $\lambda_c$, where $\text{sg}[\cdot]$ is the stopgradient operator~\cite{van2017neural}, and $a_g^\varepsilon$ is the quantized version of $a\!=\!g(x_n, \varepsilon)$. 
The former updates the anchor values to move towards the attention map $a$, and the latter forces the attention module to commit to the anchor values. 
We set $\beta\!=0.01$, $\lambda_g\!=\!0.4$, and $\lambda_c\!=\!0.1$ empirically. 
\begin{equation}\label{eq:loss_func}
\begin{split}
\mathcal{L}\!=\!-&\frac{1}{N}\sum_{n=1}^{N}\!\{ \mathbb{E}_{\epsilon\sim p(\epsilon)}[\mathbb{E}_{\varepsilon\sim p(\varepsilon)}[\log q(y_n|f(x_n, a_q^\varepsilon, \epsilon))]] \\
+ &\mathbb{E}_{\varepsilon\sim p(\varepsilon)}[\beta D_{\text{KL}}[p(z|x_n, a_q^\varepsilon) \parallel r(z)] \\
+ \lambda_q &\Vert \text{sg}[g(x_n, \varepsilon)]\!-\!a_q^\varepsilon\Vert_2^2\!+\!\lambda_c \Vert g(x_n, \varepsilon)\!-\!\text{sg}[a_q^\varepsilon] \Vert_2^2]\}.
\end{split}
\end{equation}

An illustration of the whole framework is shown in Fig.~\ref{fig:framework}. 
In practice, we first use an feature extractor, \eg VGGNet~\cite{simonyan2014very}, to extract an intermediate feature $f$ from the input $x$, then learn the attention $a, a_q$ and the variational encoding $z$ from $f$ instead of $x$.

\begin{table*}[t]
\begin{center}

\begin{threeparttable}
    \resizebox{0.95\textwidth}{!}{
    \renewcommand\arraystretch{0.95}
    \begin{tabular}{l|| c| c c c c| c c c c|| c| c | c }
    \hline\thickhline

    \multirow{2}{*}{Model} &\multirow{2}{*}{Spatial} 
          &\multicolumn{4}{c|}{CIFAR-10} &\multicolumn{4}{c||}{CIFAR-100} 
          &\multirow{2}{*}{Frequency} 
          &\multirow{2}{*}{CIFAR-10} &\multirow{2}{*}{CIFAR-100}\\
    \cline{3-10}
    & &$p\!=\!4$ &$p\!=\!8$ &$p\!=\!12$ &$p\!=\!16$ 
      &$p\!=\!4$ &$p\!=\!8$ &$p\!=\!12$ &$p\!=\!16$
    & & &\\

    \hline
    
    \hline 
    \multirow{2}{*}{VGG-att3$^*$} 
        &color &91.46 &79.61 &37.71 &~~7.12
               &52.92 &39.93 &25.40 &14.56  
        &$r\!>\!4$  &83.06 &~~3.69  \\ 
        &svhn  &90.97 &75.70 &36.74 &~~6.51  
               &82.08 &63.16 &39.07 &21.03 
        &$r\!<\!12$  &49.61 &50.90  \\ 

    \hline                                
    
    \multirow{2}{*}{VGG-aib} 
        &color &99.22 &99.69 &\textbf{93.78} &\textbf{20.59}
               &98.88 &98.56 &\textbf{95.92} &\textbf{22.70} 
        &$r\!>\!4$  &99.91 &\textbf{99.78}  \\
        &svhn  &98.59 &97.52 &97.35 &72.86
               &98.73 &96.70 &96.69 &93.02
        &$r\!<\!12$  &53.26 &79.13  \\

    \hline                                
    
    \multirow{2}{*}{VGG-aib-qt} 
        &color &\textbf{99.26} &\textbf{99.79} &91.10 &18.36
               &\textbf{99.18} &\textbf{99.30} &94.59 &20.54
        &$r\!>\!4$  &\textbf{99.96} &99.60  \\
        &svhn  &\textbf{98.65} &\textbf{98.04} &\textbf{97.85} &\textbf{78.82} 
               &\textbf{99.12} &\textbf{97.64} &\textbf{97.21} &\textbf{94.79}
        &$r\!<\!12$   &\textbf{73.52} &\textbf{79.70} \\

    \hline                                
    \hline

    \multirow{2}{*}{WRN-ABN}  
        &color &90.76 &65.01 &33.89 &13.24 
               &89.74 &61.38 &30.56 &~~9.67 %
        &$r\!>\!4$  &38.14 &14.04  \\ %
        &svhn  &90.40 &68.41 &38.46 &16.55 
               &92.77 &63.67 &30.57 &~~9.47 %
        &$r\!<\!12$  &36.33 &25.43  \\
                                      
    \hline                                
    \multirow{2}{*}{WRN-aib} 
        &color &\textbf{99.95} &\textbf{93.94} &\textbf{45.17} &~~6.69 
               &99.86 &95.35 &64.27 &17.15 
        &$r\!>\!4$  &78.96 &90.44  \\
        &svhn  &\textbf{99.94} &\textbf{97.34} &\textbf{81.98} &\textbf{47.65} 
               &99.90 &97.77 &\textbf{89.37} &62.64 
        &$r\!<\!12$  &84.58 &\textbf{94.18}  \\

    \hline                                
    \multirow{2}{*}{WRN-aib-qt} 
        &color &99.84 &84.18 &28.94 &~~4.64 
               &\textbf{99.97} &\textbf{97.07} &\textbf{69.51} &\textbf{25.80} 
        &$r\!>\!4$  &72.05 &\textbf{94.13}  \\
        &svhn  &99.91 &96.05 &70.75 &28.35 
               &\textbf{99.95} &\textbf{98.52} &89.00 &\textbf{63.10} 
        &$r\!<\!12$  &76.53 &93.17  \\
    
    \hline
    \end{tabular}
}
\end{threeparttable}
\end{center}
\caption{Interpretability scores under spatial and frequency domain modification on CIFAR-10 and CIFAR-100. $p$ is the window size of the modified region. $r$ is the radius in frequency domain. See~\S\ref{sec:interp_analysis} for more details. $^*$ denotes re-implementation. Best values in \textbf{bold}.
 \label{table:interp}}

\end{table*}

\section{Experiments}
\label{sec:experiments}
 
To demonstrate the effectiveness of the proposed IB-inspired spatial attention, we conduct extensive experiments on various visual recognition tasks, including image classification~(\S\ref{sec:image_cls}), fine-grained recognition~(\S\ref{sec:fg_cls}), and cross-dataset classification~(\S\ref{sec:cd_cls}), and achieve improved performance over baseline models. 
In~\S\ref{sec:interp_analysis}, we compare the interpretability of our attention maps with those of other attention models both qualitatively and quantitatively.
We conduct an ablation study in~\S\ref{sec:ab_study}. 
More details are shown in the supplementary.

\subsection{Image Classification}
\label{sec:image_cls}
\noindent\textbf{Datasets and models.} \textit{CIFAR-10}~\cite{krizhevsky2009learning} contains $60,000$ $32\!\times\!32$ natural images of $10$ classes, which are splited into $50,000$ training and $10,000$ test images. \textit{CIFAR-100}~\cite{krizhevsky2009learning} is similar to CIFAR-10, except that it has $100$ classes. 
We extend standard architectures, VGG and wide residual network (WRN), with the proposed IB-inspired spatial attention, and train the whole framework from scratch on CIFAR-10, and CIFAR-100, respectively.
We use original input images after data augmentation (random flipping and cropping with a padding of 4 pixels).

\noindent\textbf{Results on CIFAR.}
As shown in Table~\ref{table:visual_recognition}, the proposed attention mechanism achieves noticeable performance improvement over standard architectures and existing attention mechanisms such as GAP, PAN, \textit{VGG-att}\footnote{\tiny\url{https://github.com/SaoYan/LearnToPayAttention}} and ABN\footnote{\tiny\url{https://github.com/machine-perception-robotics-group/attention_branch_network}}. 
To be specific, \textit{VGG-aib} achieves $3.49\%$ and $9.06\%$ decrease of errors over the baseline VGG model on CIFAR-10 and CIFAR-100, respectively. 
The quantized attention model \textit{VGG-aib-qt} further decreases the errors over \textit{VGG-aib} by $0.18\%$ and $0.69\%$ on the two datasets. 
Compared with other VGG-backboned attention mechanisms, ours also achieve superior classification performances. 
Similarly, \textit{WRN-aib} and \textit{WRN-aib-qt} also decrease the top-1 errors on CIFAR-10 and CIFAR-100.

\subsection{Fine-grained Recognition}
\label{sec:fg_cls}
\noindent\textbf{Datasets and models.} \textit{CUB-200-2011} (CUB)~\cite{WahCUB_200_2011} contains $5,994$ training and $5,794$ testing bird images from $200$ classes. \textit{SVHN} collects $73,257$ training, $26,032$ testing, and $531,131$ extra digit images from house numbers in street view images.
For CUB, we perform the same preprocessing as~\cite{jetley2018learn}.
For SVHN, we apply the same data augmentation as CIFAR. 

\noindent\textbf{Results on CUB.}
The proposed \textit{VGG-aib} and \textit{VGG-aib-qt} achieves smaller classification error compared with baselines built on VGG, including GAP, PAN, and \textit{VGG-att}. 
Specially, \textit{VGG-aib} and \textit{VGG-aib-qt} outperform \textit{VGG-att} for a $3.07\%$ and $4.97\%$, respectively.

\noindent\textbf{Results on SVHN.} 
Our proposed method achieves lower errors with VGG-backboned models, and comparative errors for WRN backbone methods.

\subsection{Cross-domain Classification}
\label{sec:cd_cls}
\noindent\textbf{Datasets and models.} \textit{STL-10} contains $5,000$ training and $8,000$ test images of resolution $96\!\times\!96$ organized into $10$ classes. Following~\cite{jetley2018learn}, the models trained on CIFAR-10 are directly tested on STL-10 training and test sets without finetuning to verify the generalization abilities.

\noindent\textbf{Results on STL-10.}
As shown in Table~\ref{table:visual_recognition}, the proposed attention model outperforms other competitors on VGG backbone, and achieves comparative performance for WRN backbone.

\subsection{Interpretability Analysis}
\label{sec:interp_analysis}

An \textit{interpretable} attention map is expected to faithfully highlight the regions responsible for the model decision. 
Thus, an \textit{interpretable} attention mechanism is expected to yield consistent attention maps for an original input and a modified input if the modification does not alter the model decision.
Here, we quantify the interpretability of an attention mechanism by calculating the `interpretability score', \ie~the percentage of attention-consistent samples in prediction-consistent samples under modifications in the spatial or frequency domain. 
Here, the attention consistency is measured by the cosine similarity between two flattened and normalized attention maps.

\begin{figure}[t]
  \centering
    \includegraphics[width=0.45\textwidth]{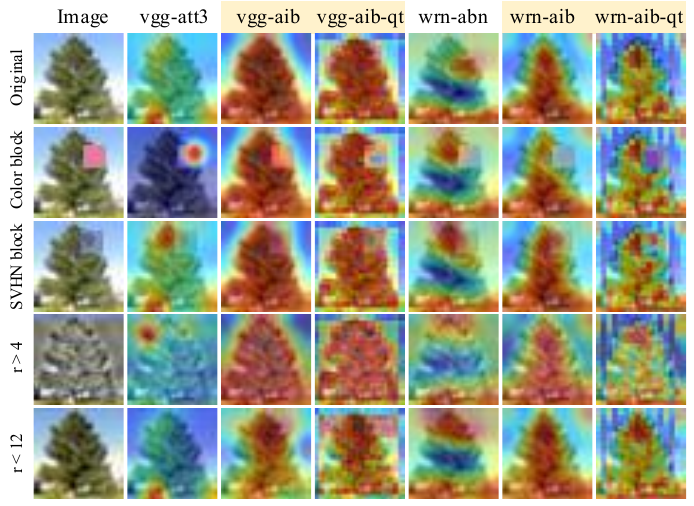}

\captionsetup{font=small}
\caption{\small{Visualization of attention maps for interpretability (\S\ref{sec:interp_analysis}).}
}
\label{fig:vis_interp}
\end{figure}

The \textit{spatial domain modification} includes randomly occluding the original images in CIFARs with color blocks or images drawn from distinct datasets. The size of the modified region $p$ ranges from $4$ to $16$. 
The $2$nd and $3$rd rows in Fig.~\ref{fig:vis_interp} show exemplar images occluded by a random color block and an image randomly drawn from SVHN, respectively, where $p\!=\!8$. As can be observed, our spatial attention model \textit{VGG-aib(-qt)}, and \textit{WRN-aib(-qt)} yield consistent attention maps with the original images (first row). 
The interpretability scores are listed in Table~\ref{table:interp}. Our method consistently outperforms other spatial attention mechanisms with the same backbone by a large margin on two datasets for various $p$. 
This is because the IB-inspired attention mechanism minimizes the MI between the attention-modulated features and the inputs, thus mitigating the influence of ambiguous information exerted on the inputs to some extent.

We also conduct \textit{frequency domain modification}, which is done by performing Fourier transform on the original samples, and feeding only high-/low-frequency (HF/LF) components into the model. 
To preserve enough information and maintain the classification performance, we focus on HF components of $r\!>\!4$, and LF components of $r\!<\!12$, where $r$ is the radius in frequency domain defined as in~\cite{wang2020high}. 
The $4$th and $5$th rows in Fig.~\ref{fig:vis_interp} show exemplar images constructed from HF and LF components, respectively. Our attention maps are more royal to those of the original images compared with other competitors. 
Out method also yields much better quantitative results than other attention models, as illustrated in Table~\ref{table:interp}. 
This is because our IB-inspired attention mechanism can well capture the task-specific information in the input by maximizing the MI between the attention-modulated feature and the task output, even when part of the input information is missing.

\subsection{Ablation Study}
\label{sec:ab_study}
We conduct ablation studies on \textit{CIFAR-10} with the VGG backbone to assess the effectiveness of each component.

\noindent\textbf{Effectiveness of attention module.} Table~\ref{table:visual_recognition} shows that \textit{VGG-aib} and \textit{VGG-aib-qt} outperform the IB-based representation learning model \textit{VGG-DVIB}.

\begin{figure}[t]
  \centering
    \includegraphics[width=\columnwidth]{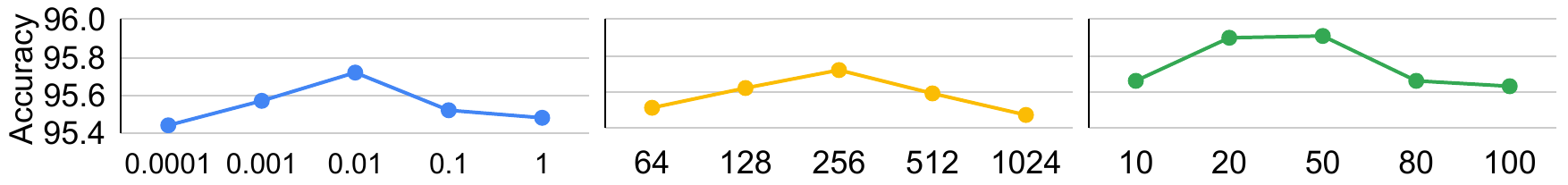}
    \put(-192,-6){\fontsize{7pt}{7pt}\selectfont{$\beta$}}
    \put(-116,-6){\fontsize{7pt}{7pt}\selectfont{$K$}}
    \put(-40,-6){\fontsize{7pt}{7pt}\selectfont{$Q$}}
    \put(-170,20){\fontsize{8pt}{8pt}\selectfont{(a)}}
    \put(-94,20){\fontsize{8pt}{8pt}\selectfont{(b)}}
    \put(-18,20){\fontsize{8pt}{8pt}\selectfont{(c)}}

\captionsetup{font=small}
\caption{\small{\textbf{Ablation study} on CIFAR-10 with VGG backbone (\S\ref{sec:ab_study}).}
}
\label{fig:ablation_study}
\end{figure}

\noindent\textbf{Information tradeoff.} $\beta$ controls the amount of information flow that bypasses the bottleneck of the network. To measure the influence of $\beta$ on the performance, we plot the accuracy values with varying $\beta$ in Fig.~\ref{fig:ablation_study} (a). As can be observed, the accuracy is largest when $\beta\!=\!0.01$. 

\noindent\textbf{Latent vector dimension.} We experiment on $K\!=\!64,128$, $256,512$, $1024$.
As shown in Fig.~\ref{fig:ablation_study} (b), $K\!=\!256$ achieves the best performance.

\begin{figure}[t]
  \centering
    \includegraphics[width=0.4\textwidth]{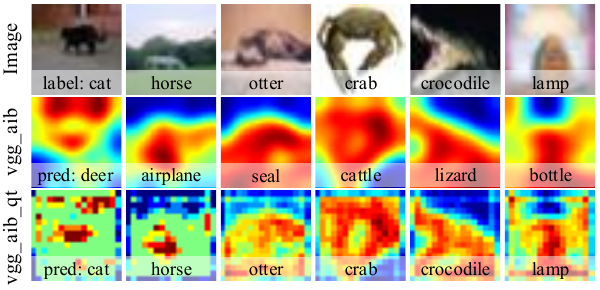}

\captionsetup{font=small}
\caption{\small{Effect of attention score quantization. See~\S\ref{sec:ab_study} for details.}
}
\label{fig:att_maps}
\end{figure}

\noindent\textbf{Attention score quantization.} Fig.~\ref{fig:ablation_study} (c) shows the classification accuracy when varying number of anchor values $Q$, where $Q$ between $20$ and $50$ gives better performance. 
Exemplary attention maps of cases that are wrongly classified by \textit{VGG-aib} but are correctly predicted by \textit{VGG-aib-qt} are listed in Fig.~\ref{fig:att_maps}. As can be observed, attention quantization can further help focus on more concentrated regions with important information, thus improve the prediction accuracy. 

\section{Conclusion}
\label{sec:conclusion}

We propose a spatial attention mechanism based on IB theory for generating probabilistic maps that minimizes the MI between the masked representation and the input, while maximizing the MI between the masked representation and the task label. 
To further restrict the information bypassed by the attention map, we incorporate an adaptive quantization mechanism to regularize the attention scores by rounding the continuous scores to the nearest anchor value during training. 
Extensive experiments show that the proposed IB-inspired spatial attention mechanism significantly improves the performance for visual recognition tasks by focusing on the most informative parts of the input. 
The generated attention maps are interpretable for the decision making of the DNNs, as they consistently highlight the informative regions for the original and modified inputs with the same semantics.

\section*{Acknowledgments}
\label{sec:acknowledgments}
This work is supported in part by General Research Fund (GRF) of Hong Kong Research Grants Council (RGC) under Grant No. 14205018 and No. 14205420.

\newpage
\bibliographystyle{named}
\bibliography{ijcai21}





\onecolumn

\begin{appendices}

In this supplementary, we first present the detailed deriviation of the proposed spatial attention mechanism inspired by the Information Bottleneck (IB)~\citeAppendix{tishby1999information} principle. We then provide explanation of the recognition tasks in our experiments. Finally, we show the implementation details of our model. 

\section{Deriviation of the IB-inspired Spatial Attention}
\label{sec:deduction}


In this section, we present the detailed deriviation of the variational lowerbound of the attentive variational information bottleneck $\mathcal{L}_{\text{AttVIB}}\!\equiv\!I(Z; Y)\!-\!\beta I(Z;X, A)$.

\begin{figure}[h]
  \centering
    \includegraphics[width=0.1\textwidth]{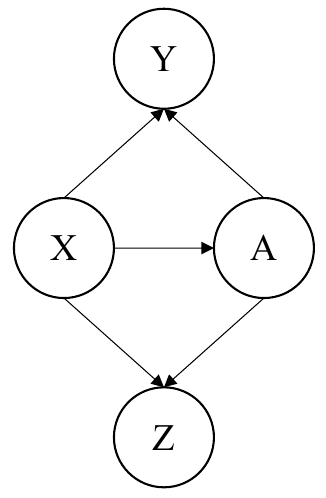}

\captionsetup{font=small}
\caption{\small{\textbf{Graphical model} of the probabilistic neural network with IB-inspired spatial attention mechanism.}
}
\label{fig:graph}
\end{figure}

\subsection{Lowerbound of $I(Z; Y)$}
\label{sec:IZY_bound}

Let the random variables $X, Y, A$ and $Z$ denote the input, output, the spatial attention map, and the latent representation obtained from $X$ and $A$, respectively. The MI between the latent feature $Z$ and its output $Y$ is defined as:
\begin{equation}\label{eq:ZY_lb}
I(Z; Y)= D_{\text{KL}}[p(y, z)||p(y)p(z)] 
   = \int p(y, z) \log \frac{p(y, z)}{p(y)p(z)} dydz 
     = \int p(y, z) \log \frac{p(y|z)}{p(y)} dydz.
\end{equation}
We introduce $q(y|z)$ to be a variational approximation of the intractable $p(y|z)$. Since the Kullback Leibler (KL) divergence is always non-negative, we have:
\begin{equation}\label{eq:KL_nonneg}
D_{KL}[p(Y|Z)||q(Y|Z)]=\int p(y|z)\log \frac{p(y|z)}{q(y|z)} dy \ge 0 
        \Rightarrow \int p(y|z)\log p(y|z) dy \ge \int p(y|z)\log q(y|z) dy, 
\end{equation}
which leads to
\begin{equation}\label{eq:ZY_lb}
I(Z; Y)\!\ge\! \int p(y, z)\! \log \frac{q(y|z)}{p(y)} dydz 
     \!=\! \int p(y, z)\! \log q(y|z)dydz\! -\! \int p(y) \log p(y) dy
     \!=\! \int p(y, z)\! \log q(y|z)dydz\! +\! H(Y),
\end{equation}
where $H(Y)\!=\!-\int p(y) \log p(y) dy$ is the entropy of $Y$, which is independent of the optimization procedure thus can be ignored. 
By leveraging the fact that $p(y, z)\!=\!\int p(x, y, z, a)dxda\!=\!\int p(x)p(a|x)p(y|x, a)p(z|x, a)dxda$ (see Fig.~\ref{fig:graph}), and ignoring $H(Y)$, 
the new variational lower bound is as follows:
\begin{equation}\label{eq:ZY_lb_two}
  \begin{split}
    I(Z; Y)&\ge \int p(x)p(a|x)p(y|x, a)p(z|x, a) \log q(y|z) dxdadydz \\
      &=\int p(x, a)p(y|x, a)p(z|x, a) \log q(y|z) dxdadydz \\
      &=\int p(x, y, a)p(z|x, a) \log q(y|z) dxdadydz \\
      &=\int p(x, y)p(a|x, y)p(z|x, a) \log q(y|z) dxdadydz.
  \end{split}
\end{equation}
Since $A$ only depends on $X$ (\cf~Fig.~\ref{fig:graph}), $p(a|x, y)$ can be simplified as $p(a|x)$. Thus Eq.~(\ref{eq:ZY_lb_two}) can be written as:
\begin{equation}\label{eq:ZY_lb_two_cont}
      I(Z; Y)\ge\int p(x, y) p(a|x) p(z|x, a) \log q(y|z) dxdadydz,
\end{equation}
where $p(a|x)$ is the attention module, $p(z|x, a)$ is the encoder, and $q(y|z)$ is the decoder where $z$ is the output of $p(z|x, a)$. $I(Z; Y)$ can thus be maximized by maximizing its variational lower bound.

\subsection{Upperbound of $I(Z;X, A)$}
\label{sec:IZXA_bound}

Next, to minimize the MI between the masked repsentation and the input, we consider $I(Z; X, A)$, and obtain the following upper bound according to the \textbf{chain rule for mutual information}, which is calculated as follows: 
\begin{equation}\label{eq:ZXA_ub}
\begin{split}
I(Z; X, A) = &  I(Z; X) + I(Z; A | X)
     = \int p(x, z) \log \frac{p(x, z)}{p(x)p(z)} dxdz 
     + \int p(x, a, z) \log \frac{p(x)p(x, a, z)}{p(x, a)p(x, z)} dxdadz \\
     \le &\int p(x, z) \log \frac{p(z|x)}{r(z)}  dxdz 
     + \int p(x, a, z) \log \frac{p(z|x, a)}{p(z|x)} dxdadz
\end{split}
\end{equation}

where $r(z)$ is a prior distribution of $z$. This inequation is also from the nonnegativity of the Kullback Leibler (KL) divergence. 

Eq.~(\ref{eq:ZXA_ub}) can be further simplified as follows:
\begin{equation}\label{eq:ZXA_ub_simp}
\begin{split}
& I(Z; X, A) \le \int p(x, z) \log \frac{p(z|x)}{r(z)}  dxdz 
+ \int p(x, a, z) \log \frac{p(z|x, a)}{p(z|x)} dxdadz \\
=&\int p(x)p(z|x) \log \frac{p(z|x)}{r(z)} dxdz 
+ \int p(x)p(a|x)p(z|x, a) \log \frac{p(z|x, a)}{p(z|x)} dxdadz \\
= &\int p(x)p(z|x) (\int p(a|z, x) da) \log \frac{p(z|x)}{r(z)} dxdz 
+ \int p(x)p(a|x)p(z|x, a) \log \frac{p(z|x, a)}{p(z|x)} dxdadz\\
= &\int p(x)p(a, z|x)  \log \frac{p(z|x)}{r(z)} dxdadz 
+ \int p(x)p(a|x)p(z|x, a) \log \frac{p(z|x, a)}{p(z|x)} dxdadz\\
= &\int p(x)p(a|x)p(z|x, a) \log \left(\frac{p(z|x, a)}{p(z|x)} \frac{p(z|x)}{r(z)}\right) dxdadz 
= \int p(x)p(a|x)p(z|x, a) \log \frac{p(z|x, a)}{r(z)} dadz
\end{split}
\end{equation}

By combining the two bounds in Eq.~(\ref{eq:ZY_lb_two}) and~(\ref{eq:ZXA_ub_simp}), we obtain the attentive variational information bottleneck:
\begin{equation}\label{eq:IB_two}
\begin{split}
\mathcal{L}_{\text{AttVIB}} &\equiv I(Z; Y) - \beta I(Z;X, A) \\
= \int &p(x, y) p(a|x) p(z|x, a) \log q(y|z) dxdadydz 
  - \beta \int p(x)p(a|x)p(z|x, a) \log \frac{p(z|x, a)}{r(z)} dxdadz,
\end{split}
\end{equation}
which offers an IB-inspired spatial attention mechanism for visual recognition.

Following~\citeAppendix{alemi2017deep}, we approximate $p(x)$ and $p(x, y)$ with empirical data distribution $p(x) = \frac{1}{N}\sum_{n=1}^{N}\delta_{x_n}(x)$ and $p(x, y) = \frac{1}{N}\sum_{n=1}^{N}\delta_{x_n}(x)\delta_{y_n}(y)$, where $N$ is the number of training samples, $x_n$ and $(x_n, y_n)$ are samples drawn from data distribution $p(x)$ and $p(x, y)$, respectively. 
The approximated lower bound $\mathcal{L}_{\text{AttVIB}}$ can thus be written as:
\begin{equation}\label{eq:L_IB}
\widetilde{\mathcal{L}}_{\text{AttVIB}} = \frac{1}{N}\sum_{n=1}^{N}  \{\int p(a|x_n) p(z|x_n, a) \log q(y_n|z) dadz 
        -\beta \int p(a|x_n)p(z|x_n, a) \log \frac{p(z|x_n, a)}{r(z)} dadz\}.
\end{equation}
Similar to~\citeAppendix{alemi2017deep}, we suppose to use the attention module of the form $p(a|x)\!=\!\mathcal{N}(a| g_e^{\mu}(x), g_e^{\Sigma}(x))$, and the encoder of the form $p(z| x, a)\!=\!\mathcal{N}(z| f_e^{\mu}(x, a), f_e^{\Sigma}(x, a))$, where $g_e$ and $f_e$ are network modules. 
To enable back-propagation, we use the reparametrization trick~\citeAppendix{kingma2014auto}, and write $p(a|x)da\!=\!p(\varepsilon)d\varepsilon$ and $p(z|x, a)dz\!=\!p(\epsilon)d\epsilon$, where $a\!=\!g(x, \varepsilon)$ and $z\!=\!f(x, a, \epsilon)$ are deterministic functions of $x$ and the Gaussian random variables $\varepsilon$ and $\epsilon$. 
The final loss function is:
\begin{equation}\label{eq:L_IB_RP}
\mathcal{L}\!=\!-\frac{1}{N}\sum_{n=1}^{N}\!\{ \mathbb{E}_{\epsilon\sim p(\epsilon)}[\mathbb{E}_{\varepsilon\sim p(\varepsilon)}[\log q(y_n|f(x_n, g(x_n, \varepsilon), \epsilon))]] 
+ \beta \mathbb{E}_{\varepsilon\sim p(\varepsilon)}[D_{KL}[p(z|x_n, g(x_n, \varepsilon)) \parallel r(z)]]\}.
\end{equation}
The first term is the negative log-likelihood of the prediction, where the label $y_n$ of $x_n$ is predicted from the latent encoding $z$, and $z$ is generated from the $x_n$ and the attention $a=g(x_n, \varepsilon)$. Minimizing this term leads to maximal prediction accuracy. 
The second term is the KL divergence between distributions of latent encoding $z$ and the prior $r(z)$. 
The minimization of the KL terms in Eq.~(\ref{eq:L_IB_RP}) enables the model to learn an IB-inspired spatial attention mechanism for visual recognition.

\section{Introduction of the recognition tasks}
\label{sec:task_inro}


\noindent\textbf{Fine-grained recognition} is to differentiate similar subordinate categories of a basic-level category, \eg~bird species~\citeAppendix{WahCUB_200_2011}, flower types~\citeAppendix{nilsback2008automated}, dog breeds~\citeAppendix{khosla2011novel}, automobile models~\citeAppendix{krause20133d}, \etc~ 
The annotations are usually provided by domain experts who are able to distinguish the subtle differences among the highly-confused sub-categories. 
Compared with generic image classification task, fine-grained recognition would benefit more from finding informative regions that highlight the visual differences among the subordinate categories, and extracting discriminative features therein.

\noindent\textbf{Cross-domain classification} is to test the generalization ability of a trained classifier on other domain-shifted benchmarks, \ie~datasets that have not been used during training. 
In our experiments, we directly evaluate the models trained on CIFAR-10 on the \texttt{train} and \texttt{test} set of STL-10 following~\citeAppendix{jetley2018learn}.

\section{Implementation details}
\label{sec:impl_details}

\noindent\textbf{Network Configuration.} 
For CIFAR and SVHN, the feature extractor consists of the first three convolutional blocks of VGG, where the max-pooling layers are omitted, or the first two residual blocks of WRN. 
In \textit{VGG-aib}, $g_e^{\mu}$ for obtaining the $\mu_a$ is built as: \textit{Conv}($3\!\times\!3, 1$) $\rightarrow$ \textit{Sigmoid}, and $\sigma_a$ is calculated from $\mu_a$ through a \textit{Conv}($1\!\times\!1, 1$), which is inspired by~\citeAppendix{bahuleyan2018variational}. 
In \textit{WRN-aib}, $g_e^{\mu}$ contains an extra residual block compared with that in VGG backbone, and $\sigma_a$ is obtained in the same way. 
The encoder contains the remaining convolutional building blocks from VGG or WRN, with an extra \textit{FC} layer that maps the flattened attention-modulated feature to a $2K$ vector, where the first $K$-dim is $\mu$, and the remaining $K$-dim is $\sigma$ (after a softplus activation).
The decoder maps the sampled latent representation $z$ to the prediction $y$, which is built as: \textit{FC}($K\!\times\!K$) $\rightarrow$\textit{ReLU}$\rightarrow$\textit{FC}($K\!\times\!C$) for VGG backbone, and \textit{ReLU}$\rightarrow$\textit{FC}($K\!\times\!C$) for WRN backbone, where $K$ is the dimension of the latent encoding, and $C$ is the number of classes. 
For CUB, the feature extractor is built from the whole convolutional part of VGG. The encoder only contains a Linear layer. The structure of the attention module and the decoder is the same as above.
For \textit{VGG-aib-qt} and \textit{WRN-aib-qt}, the number of anchor values $Q$ are $20$ and $50$, respectively. The values $\{v_i\}$ are initialized by evenly dividing $[0,1]$ and then trained end-to-end within the framework.

\noindent\textbf{Comparions of \#parameters.}
In Table.~\ref{table:num_parameters}, we report the number of parameters for different attention mechanisms under various settings. We will add this to the future version.

\begin{table*}[t]
  \begin{center}
  \caption{\small Comparisons of the number of parameters (M).}
  \label{table:num_parameters}
  \begin{threeparttable}
  \resizebox{0.65\textwidth}{!}{
  \setlength\tabcolsep{1.5pt}
  \renewcommand\arraystretch{1.2}
  \begin{tabular}{c c||c c c c c c}
  \hline\thickhline
  Input Size &\#Class  & VGG-att3 &VGG-aib &VGG-aib-qt & WRN-ABN &WRN-aib &WRN-aib-qt \\
  \hline
  ~$32\!\times\!32$, &10 &19.86 &20.05 &20.05 &64.35 &~~65.00 &~~65.00 \\
  ~~$32\!\times\!32$, &100 &19.99 &20.07 &20.07 &64.48 &~~65.11 &~~65.11 \\
  $224\!\times\!224$, &200  &- &24.28 &24.28 &- &130.73 &130.73 \\

  \hline
  \end{tabular}
  }
  \end{threeparttable}
  \end{center}
\end{table*}

\noindent\textbf{Hyper-parameter selection.} 
We first determine the ranges of the hyper-parameters by referring to related works~\citeAppendix{alemi2017deep,van2017neural}, and then discover the best ones through experiments. 
Following this scheme, we choose a good $\beta$ for different backbones by experimenting on $\beta\!=\!10^{-p}, p=0,\dots,4$.
For a certain backbone, the hyper-parameters are the same for tasks whose inputs share the same resolution. The hyper-parameters varied slightly for different backbones. 
Latent vector dimension $K$ defines the size of the bottleneck as in~\citeAppendix{alemi2017deep}, which does not necessarily match the latent dimension of other architectures.

\noindent\textbf{Training.} 
Models are trained from scratch except on CUB following~\citeAppendix{jetley2018learn}. 
We use an SGD optimizer with weight decay $5\times 10^{-4}$ and momentum $0.9$, and train for $200$ epochs in total. 
The batch size is $128$ for all the datasets.
For \textit{VGG-aib(-qt)}, the initial learning rate is $0.1$, which is scaled by $0.5$ every $25$ epochs. For \textit{WRN-aib(-qt)}, the initial learning rate is $0.1$, and is multiplie,d by $0.3$ at the $60, 120$ and $180$-th epoch. 

\noindent\textbf{Inference.}  
We draw $4$ samples from $p(\varepsilon)$ for $a$ and $12$ samples from $p(\epsilon)$ for $z$ in the lower bound of the IB-inspired attention mechanism in Eq.~(\ref{eq:IB_two}) and the loss in Eq.~(\ref{eq:L_IB}). 

\noindent\textbf{Running time analysis.}
We show the FLOPs and inference time of our method in Table.~\ref{table:running_rime}. 
\begin{table*}[t]
  \begin{center}
  \caption{\small Running time analysis with various task settings.}
  \label{table:running_rime}
  \begin{threeparttable}
  \resizebox{0.8\textwidth}{!}{
  \setlength\tabcolsep{1.5pt}
  \renewcommand\arraystretch{1.2}
  \begin{tabular}{c c||c c c c ||c c c c}
  \hline\thickhline
  \multirow{2}{*}{Input Size} 
  &\multirow{2}{*}{\#Class} 
  &\multicolumn{4}{c||}{FLOPs (G)} &\multicolumn{4}{c}{Inference time per frame (ms)} \\
  \cline{3-10}
  &&VGG-aib &VGG-aib-qt  &WRN-aib &WRN-aib-qt
   &VGG-aib &VGG-aib-qt  &WRN-aib &WRN-aib-qt\\
  \hline
  ~$32\!\times\!32$ &10 
          &~~3.79 &~~3.79 &39.14 &39.14 
          &3.43  &3.68  &~~8.53  &~~8.55 \\
  ~~$32\!\times\!32$ &100 
          &~~3.79 &~~3.79 &39.20 &39.20 
          &3.51  &3.70  &~~8.60  &~~8.84 \\
  $224\!\times\!224$ &200 
          &16.75 &16.75 &50.97 &50.97 
          &3.58  &3.81  &10.47  &10.66 \\

  \hline
  \end{tabular}
  }
  \end{threeparttable}
  \end{center}
\end{table*}

\noindent\textbf{Reproducibility.} Our model is implemented using \textit{PyTorch} and trained/tested on one Nvidia TITAN Xp GPU. To ensure reproducibility, 
our code is released at \url{https://github.com/ashleylqx/AIB.git}.

\bibliographystyleAppendix{named}
\bibliographyAppendix{varef}

\end{appendices}

\end{document}